\def\year{2020}\relax
\documentclass[letterpaper]{article} 
\usepackage{aaai20}  
\usepackage{times}  
\usepackage{helvet} 
\usepackage{courier}  
\usepackage[hyphens]{url}  
\usepackage{graphicx} 
\usepackage{graphicx}  
\usepackage{amssymb}
\usepackage{booktabs}
\usepackage{multirow}
\usepackage{makecell}
\usepackage{amsmath}
\usepackage{amsmath,amsfonts,amssymb,bbm,epsfig,bm}
\usepackage{algorithm}
\usepackage{algorithmicx}
\usepackage{algpseudocode}
\newcommand{\citet}[1]{\citeauthor{#1}~\shortcite{#1}}
\usepackage{multirow}

\title{Graph-Based Reasoning over Heterogeneous External  Knowledge for  Commonsense Question Answering}

\author{Shangwen Lv\textsuperscript{\rm 1,2}\thanks{Euqal Contributions. Work was done while this author was an intern at Microsoft Research Asia.}, Daya Guo\textsuperscript{\rm 3}\footnotemark[1], Jingjing Xu\textsuperscript{\rm 4}\footnotemark[1], Duyu Tang\textsuperscript{\rm 5}, Nan Duan\textsuperscript{\rm 5}, \\ \bf \Large \textbf{Ming Gong\textsuperscript{\rm 5}, Linjun Shou\textsuperscript{\rm 5}, Daxin Jiang\textsuperscript{\rm 5}, Guihong Cao\textsuperscript{\rm 5}, Songlin Hu\textsuperscript{\rm 1,2}}\\
\textsuperscript{\rm 1}Institute of Information Engineering, Chinese Academy of Sciences, Beijing, China\\
\textsuperscript{\rm 2}School of Cyber Security, University of Chinese Academy of Sciences, Beijing, China\\
\textsuperscript{\rm 3}Sun Yat-sen University \
\textsuperscript{\rm 4}Peking University \
\textsuperscript{\rm 5} Microsoft Corporation \\
\{lvshangwen, husonglin\}@iie.ac.cn \\
guody5@mail2.sysu.edu.cn,
jingjingxu@pku.edu.cn\\
\{dutang,nanduan,migon,lisho,djiang,gucao\}@microsoft.com \\
}

\begin{document}
\maketitle
\begin{abstract}
Commonsense  question answering aims to answer questions which require background knowledge that is not explicitly expressed in the question. The key challenge is how to obtain evidence from external knowledge and make predictions based on the evidence. Recent studies either learn to generate evidence from human-annotated evidence which is expensive to collect, or extract evidence from either structured or unstructured knowledge bases which fails to take advantages of both sources simultaneously. In this work, we propose to automatically extract evidence from heterogeneous knowledge sources, and answer questions based on the extracted evidence. Specifically, we extract evidence from both structured knowledge base (i.e. ConceptNet) and Wikipedia plain texts. We construct graphs for both sources to obtain the relational structures of evidence. Based on these graphs, we propose a graph-based approach consisting of a graph-based contextual word representation learning module and a graph-based inference module. The first module utilizes graph structural information to re-define the distance between words for learning better contextual  word representations. The second module adopts graph convolutional network to encode neighbor information into the representations of nodes, and aggregates evidence with graph attention mechanism for predicting the final answer. Experimental results on CommonsenseQA dataset illustrate that our graph-based approach over both knowledge sources brings improvement over strong baselines. Our approach achieves the state-of-the-art accuracy (75.3\%) on the CommonsenseQA dataset.

\end{abstract}

\section{Introduction}

Reasoning is an important and challenging task in artificial intelligence and natural language processing, which is ``\textit{the process of drawing conclusions from the principles and evidence}''  \cite{wason1972psychology}. 
The ``\textit{evidence}'' is the fuel and the ``\textit{principle}'' is the machine that operates on the fuel to make predictions.
The majority of studies typically only take the current datapoint as the input, in which case the important ``\textit{evidence}'' of the datapoint from background knowledge is ignored.

\begin{figure}[t]
	\centering
	\includegraphics[width=0.45\textwidth]{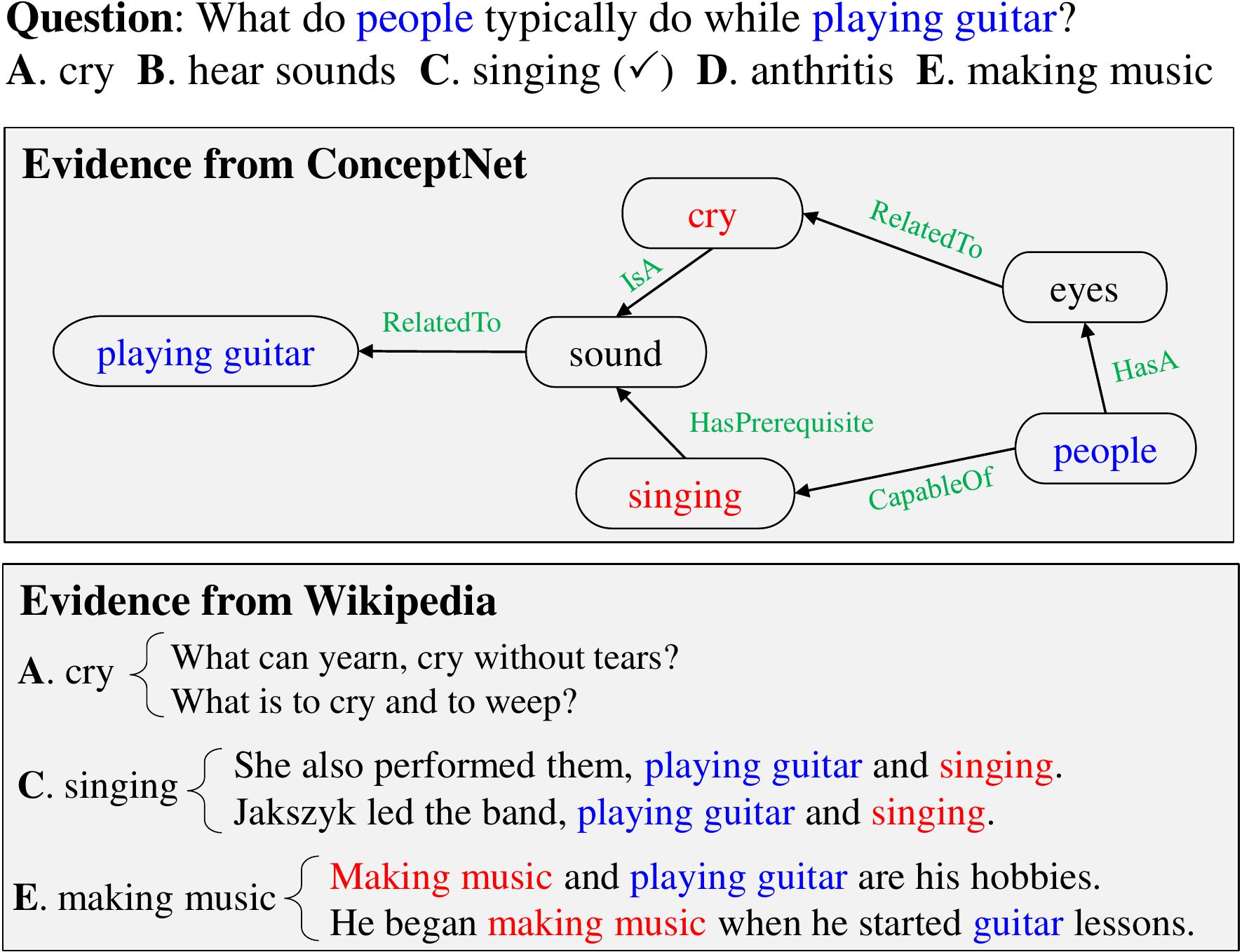}
	\caption{An example from the CommonsenseQA dataset which requires multiple external knowledge to make the correct prediction. ConceptNet evidence helps pick up choices (A, C) and Wikipedia evidence helps pick up choices (C, E). Combining both evidence will derive the right answer C. Words in blue are the concepts in the question. Words in green are the relations from ConceptNet. Words in red are the choices picked up by evidence.}
	\label{fig:intro_example}
\end{figure}

In this work, we study commonsense question answering, a challenging task which requires machines to collect background knowledge and reason over the knowledge to answer questions. For example, an influential dataset CommonsenseQA \cite{talmor2019commonsenseqa} is built in a way that the answer choices share the same relation with the concept in the question while annotators are asked to use their background knowledge to create questions so that only one choice is the correct answer. Figure~\ref{fig:intro_example} shows an example which requires multiple external knowledge sources to make the correct predictions. 
The structured evidence from ConcepNet can help pick up the choices (A, C), while evidence from Wikipedia can help pick up the choices (C, E).
Combining both evidence will derive the correct answer (C).

Approaches have been proposed in recent years for extracting evidence and reasoning over evidence. Typically, they either generate evidence from human-annotated evidence~\cite{RajaniMXS19} or extract evidence from a homogeneous knowledge source like structured knowledge ConceptNet \cite{kag2019,bauer2018commonsense,mihaylov2018knowledgeable} or Wikipedia plain texts \cite{ryu2014open,yang2015wikiqa,chen2017reading}, but they fail to take  advantages of both knowledge sources simultaneously. Structured knowledge sources contain valuable structural relations between concepts, which are beneficial for reasoning. However, they suffer from low coverage. Plain texts can provide abundant and high-coverage evidence, which is complementary to the structured knowledge.

In this work, we study commonsense question answering by 
using automatically 
collected evidence from heterogeneous external knowledge.
Our approach consists of two parts: knowledge extraction and graph-based reasoning. In the knowledge extraction part, we automatically extract graph paths from ConceptNet and sentences from Wikipedia. 
To better use the relational structure of the evidence, we construct graphs for both sources, including extracted graph paths from ConceptNet and 
triples derived from Wikipedia sentences by Semantic Role Labeling (SRL).
In the graph-based reasoning part, we propose a graph-based approach to make better use of the graph information.
We contribute by developing two graph-based modules, including (1) a graph-based contextual word representation learning module, which  utilizes graph structural information to re-define the distance between words for learning better contextual word representations, and (2) a graph-based inference module, which first
adopts Graph Convolutional Network \cite{kipf2016semi} to encode neighbor information into the representations of nodes, followed by a graph attention mechanism for evidence aggregation.

We conduct experiments on the CommonsenseQA benchmark dataset. 
Results show that both the graph-based contextual representation learning module and the graph-based inference module boost the performance.
We also demonstrate that incorporating both knowledge sources can bring further improvements.
Our approach
achieves the state-of-the-art accuracy (75.3\%) on the CommonsenseQA dataset.

Our contributions of this paper can be summarized as follows:
\begin{itemize}
    \item  We introduce a graph-based approach to leverage evidence from heterogeneous knowledge sources 
    for commonsense question answering.
    \item We propose a graph-based contextual representation learning module and a graph-based inference module to make better use of the graph information for commonsense question answering.
    \item Results show that our model achieves a new state-of-the-art performance on the CommonsenseQA dataset.
\end{itemize}

 \section{Task Definition and Dataset}
 This paper utilizes CommonsenseQA \cite{talmor2019commonsenseqa}, an influential dataset for commonsense question answering task for experiments. Formally, given a natural language question $Q$ containing $m$ tokens $\{ q_1, q_2, \cdots, q_m\}$, and $5$ choices $\{ a_1, a_2, \cdots, a_5\}$, the target is to distinguish the right answer from the wrong ones and accuracy is adopted as the metric. 
 Annotators are required to utilize their background knowledge to write questions in which only one of them is correct, thus making the task more challenging. 
The lack of evidence requires the model to have strong commonsense knowledge extraction and reasoning ability to get the right results.

\section{Approach Overview}
\begin{figure}[H]
	\centering
	\includegraphics[width=0.45\textwidth]{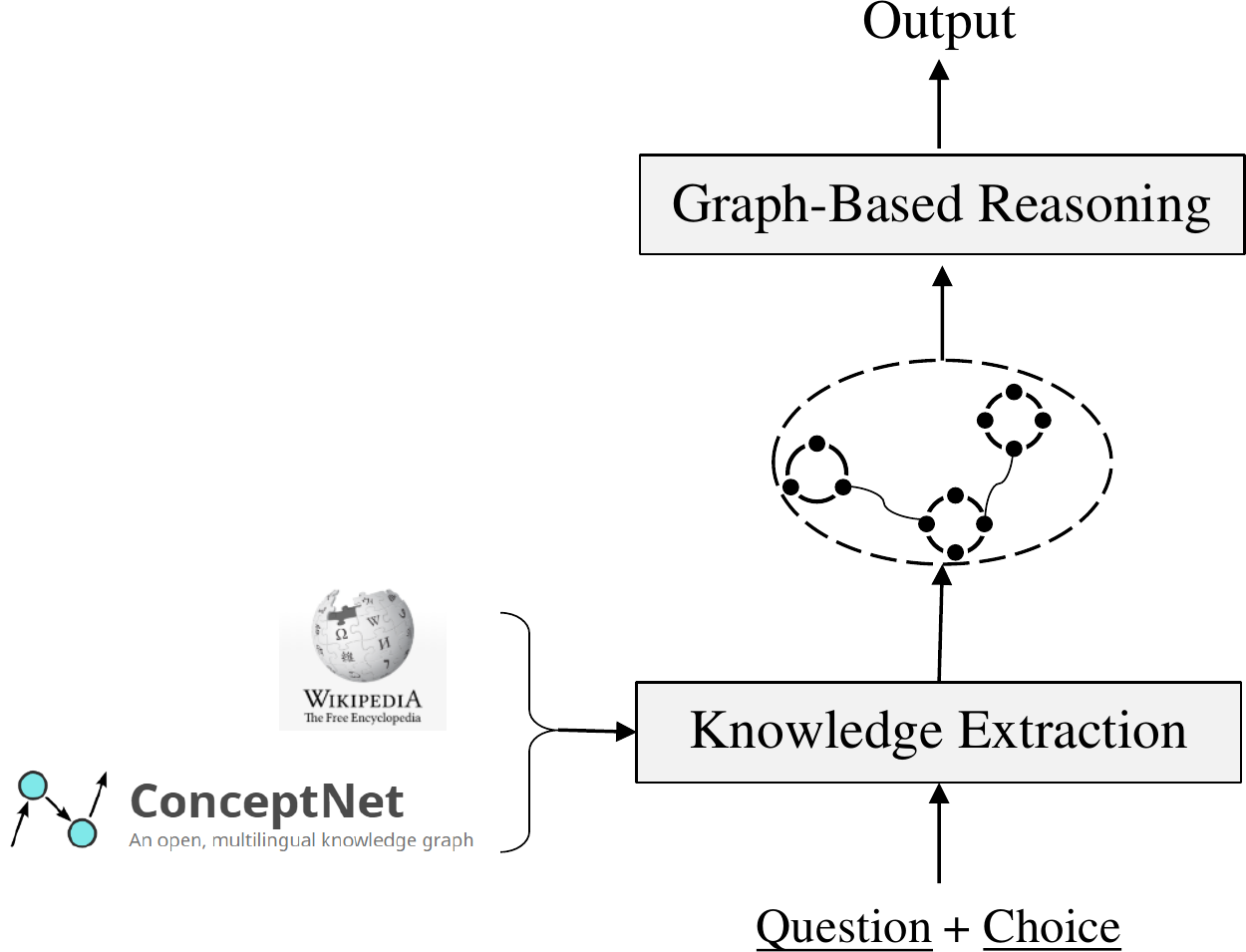}
	\caption{An overview of our approach.}
	\label{fig:overview}
\end{figure}

In this section, we give an overview of our approach. As shown in Figure \ref{fig:overview}, our approach contains two parts: knowledge extraction and graph-based reasoning. In the knowledge extraction part, we extract knowledge from structured knowledge base ConcpetNet  and Wikipedia plain texts according to the given question and choices. We construct graphs to utilize the relational structures of both sources. 
In the graph-based reasoning part, we propose two graph-based modules which consists of a graph-based contextual word representation learning module and a graph-based inference module to infer final answers. We will describe each part in detail in the following sections.

 \section{Knowledge Extraction}

In this section, we provide the methods to extract evidence from ConceptNet and Wikipedia given the question and choices. Furthermore, we describe the details of constructing graphs for both sources.

\subsection{Knowledge Extraction from ConceptNet}
\begin{figure}[htbp]
	\centering
	\includegraphics[width=0.35\textwidth]{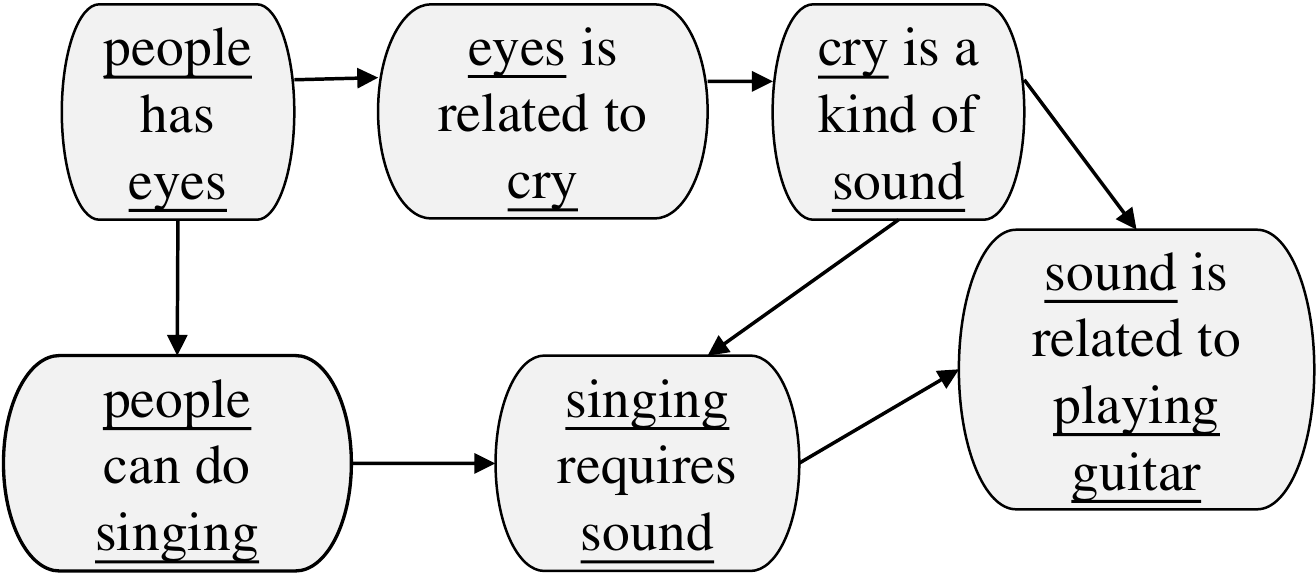}
	\caption{An example of constructed Concept-Graph from the ConceptNet evidence. The underlined words are the concepts in ConceptNet.}
	\label{fig:concept_graph}
\end{figure}

ConceptNet is a large-scale commonsense knowledge base, containing millions of nodes and relations.  The triple in ConceptNet contains four parts: two nodes, one relation, and a relation weight. 
For each question and choice, we first identify their entities in the given ConceptNet graph. Then we search for the paths (less than 3  hops) from question entities to choice entities and merge the covered triples into a graph where nodes are triples and edges are the relation between triples.  If two triples $s_i$, $s_j$ contain the same entity, we will add an edge from the previous triple $s_i$ to the next triple $s_j$. In order to obtain contextual word representations for ConceptNet nodes, we transfer the triple into a natural language sequence according to the relation template in ConceptNet. 
An example is shown in Figure~\ref{fig:concept_graph}.  We denote the graph as Concept-Graph.

\subsection{Knowledge Extraction from Wikipedia}
We extract 107M sentences from Wikipedia\footnote{Wikipedia version enwiki-20190301} by Spacy\footnote{https://spacy.io/} and adopt Elastic Search tools\footnote{https://www.elastic.co/} to index the Wikipedia sentences. We first remove stopwords in the given question and choices then concatenate the words as queries to search from the Elastic Search engine. The engine ranks the matching scores between queries and all the Wikipedia sentences. We select top $K$  sentences as the Wikipedia evidence. Here we adopt $K$=10 in experiments.

\begin{figure}[htbp]
	\centering
	\includegraphics[width=0.4\textwidth]{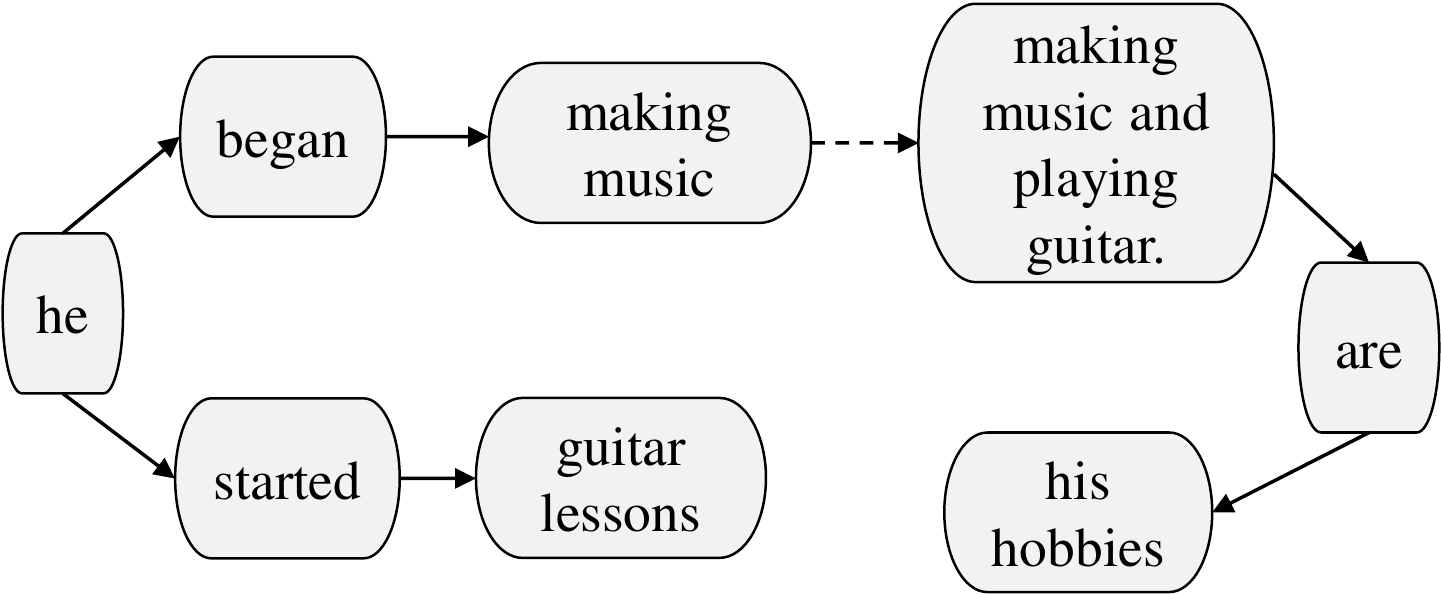}
	\caption{An example of constructed Wiki-Graph from the Wikipedia evidence ``He began making music when he started guitar lessons'' and ``Making music and playing guitar are his hobbies''. The dotted line denotes the added edge according to our enhanced rules (1).}
	\label{fig:wiki_graph}
\end{figure}

To discover the structure information in Wikipedia evidence, we construct a graph for Wikipedia evidence. We utilize Semantic Role Labeling (SRL) to extract triples (subjective, predicate, objective) in one sentence. Both arguments and predicates are the nodes in the graph. We add two edges $<$subjective, predicate$>$ and $<$predicate, objective$>$ in the graph. In order to enhance the  connectivity of the graph. We remove stopwords and add an edge from node $a$ to node $b$ according to the following enhanced rules: (1) Node $a$ is contained in node $b$ and the number of words in $a$ is more than 3;  (2) Node $a$ and node $b$ only have one different word and the numbers of words in $a$ and $b$ are both more than 3. An example is shown in Figure~\ref{fig:wiki_graph}. We denote the  graph as Wiki-Graph.

\section{Graph-Based Reasoning}
\begin{figure}[htbp]
	\centering
	\includegraphics[width=0.45\textwidth]{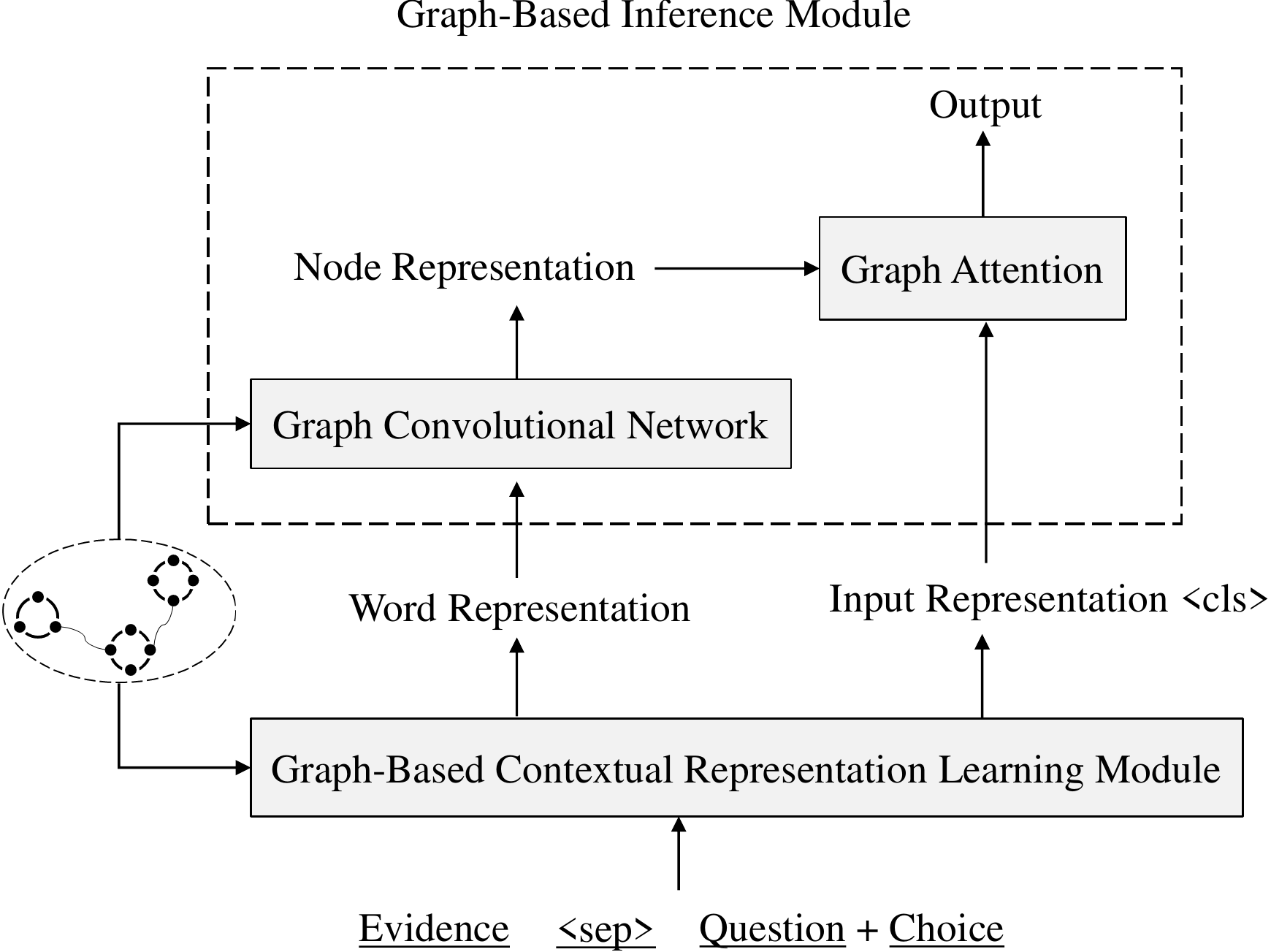}
	\caption{An overview of our proposed graph-based reasoning model.}
	\label{fig:model}
\end{figure}

In this section, we present the model architecture of graph-based reasoning over the extracted evidence, shown in Figure \ref{fig:model}. Our graph-based model consists of two modules: a graph-based contextual representation learning module and a  graph-based inference module. 
The  first  module  learns better contextual word representations by using graph information to re-define the distance between words. The second module gets node representations via Graph Convolutional Network \cite{kipf2016semi} by  using neighbor information and  aggregates graph representations to make final predictions.

\subsection{Graph-Based Contextual Representation Learning Module}

It is well accepted that pre-trained models have a strong text understanding ability and have achieved state-of-the-art results on a variety of natural language processing tasks. 
We use XLNet \cite{xlnet2019yang} as the backbone here, which is a successful pre-trained model with the advantage of capturing long-distance dependency.
A simple way to get the representation of each word is to concatenate all the evidence as a single sequence and feed the raw input into XLNet.
However, this would assign a long distance for the words mentioned in different evidence sentences, even though they are semantically related.
Therefore, we use the graph structure to re-define the relative position between evidence words. In this way, semantically related words will have shorter relative position and the internal relational structures in evidence are used to obtain better contextual word representations.

Specifically, we develop an efficient way of utilizing topology sort algorithm\footnote{We also try to re-define the relative positions between two word tokens and get a position matrix according to the token distances in the graph. However, it consumes too much memory and cannot be executed efficiently.} to re-order the input evidence according to the constructed graphs. 
For structured knowledge, ConceptNet triples are not represented as natural language. We use the relation template provided by ConceptNet to transfer a triple into a natural language text sentence.
For example, ``mammals HasA hair'' will be transferred to ``mammals has hair''. In this way, we can get a set of sentences $S_T$ based on the triples in the extracted graph. Then we can get the re-ordered evidence  for  ConceptNet $S_T'$ with the method shown in Algorithm~\ref{alg:training}. The output of Figure 3 is $<$``people has eyes'', ``eyes is related to cry'', ``people can do singing'', ``cry is a kind of sound'', ``singsing requires sound'', ``sound is related to playing guitar''$>$, which will shorten the distances between triples which are more similar to each other.
For Wikipedia sentences, we construct a sentence graph. The evidence sentences $S$ are nodes in the graph. For two sentences $s_i$ and $s_j$, if there is an edge $<$$p$, $q$$>$ in Wiki-Graph where $p$, $q$ are in $s_i$ and $s_j$ respectively, there will be an edge $<$$s_i$, $s_j$$>$ in the sentence graph. We can get a sorted evidence sequence $S'$ by the method in Algorithm~\ref{alg:training}. In Algorithm~\ref{alg:training}, the relations $R$ is a set of edges, and an edge $r$=$<$$x$,$y$$>$ means the edge from node $x$ to node $y$. The incident edges for $s_i$ represent edges from other nodes to the node $s_i$.  

Formally, the input of XLNet is the concatenation of sorted ConceptNet evidence sentences $S_T'$, sorted Wikipedia evidence sentences $S'$,  question $q$, and choice $c$. The output of XLNet is contextual word piece representations and the input representation $<$cls$>$. By transferring the extracted graph into natural language texts, we can fuse these two different heterogeneous knowledge sources into the same representation space.

\begin{algorithm}[t]
\centering
\footnotesize
\begin{algorithmic}[1]
\Require
A sequence of nodes $S = \{s_i, s_2, \cdots, s_n\}$; A set of relations  $R = \{r_1, r_2, \cdots, r_m\}$. 
\Function{dfs}{node, visited, sorted\_sequence} 
\For{each child $s_c$ in node's children}
\If{$s_c$ has no incident edges and visited[$s_c$]==0}
\State
visited[$s_c$]=1
\State
sorted\_sequence.append(0, $s_c$)
\State Remove the incident edges of $s_c$
\State DFS($s_c$, visited, sorted\_sequence)
\EndIf
\EndFor

\EndFunction
\State
sorted\_sequence = []
\State
visited = [0 for i in range(n)]
\State
S,R = to\_acyclic\_graph(S,R)
\For{each node $s_i$ in $S$}
\If{$s_i$ has no incident edges and visited[i] == 0}
\State
visited[i] = 1
\State sorted\_sequence.append($s_i$)
\State
DFS($s_i$, visited, sorted\_sequence)

\EndIf
\EndFor
\State
\Return sorted\_sequence
\end{algorithmic}
\caption{Topology Sort Algorithm.}
\label{alg:training}
\end{algorithm}

\subsection{Graph-Based Inference Module}
The XLNet-based model mentioned in the previous subsection provides effective word-level clues for making predictions.
Beyond that, the graph provides more semantic-level information of evidence at a more abstract layer, such as the subject/object of a relation.
A more desirable way is to aggregate evidence at the 
graph-level to make final predictions.

Specifically, we regard the two evidence graphs Concept-Graph and Wiki-Graph as one graph and adopt Graph Convolutional Networks (GCNs) \cite{kipf2016semi} to obtain node representations by encoding graph-structural information.

To propagate information among evidence and reason over the graph, GCNs update node representations by pooling features of their adjacent nodes. Because relational GCNs usually over-parameterize the model \cite{marcheggiani2017encoding,zhang2018graph}, we apply GCNs on the undirected graph.

The $i$-th node representation $h^{0}_i$ is obtained by averaging hidden states of the corresponding evidence in the output of XLNet and reducing dimension via a non-linear transformation: 
\begin{equation}\label{hidden_state}
h^{0}_i=\sigma(W\sum_{w_j\in s_i}\frac{1}{|s_i|}h_{w_j}) \,.
\end{equation} 
where $s_i=\{w_0,\cdots,w_t\}$ is the corresponding evidence to the $i$-th node, $h_{w_j}$ is the contextual token representation of XLNet for the token $w_j$, $W \in R^{d\times k}$ is to reduce high dimension $d$ into low dimension $k$, and $\sigma$ is an activation function.  

In order to reason over the graph, we propagate information across evidence via two steps: aggregation and combination \cite{hamilton2017inductive}. The first step aggregates information from neighbors of each node. The aggregated information $z^{l}_i$ for $i$-th node can be formulated as Equation \ref{aggregation}, where $N_i$ is the neighbors of $i$-th node and $h^{l}_j$ is the $j$-th node representation at the layer $l$. The representation $z^{l}_i$ contains neighbors information for $i$-th node at the layer $l$, and we can combine it with the transformed $i$-th node representation to get the updated node representation $h^{l+1}_i$:
\begin{equation}\label{aggregation}
z^{l}_i=\sum_{j\in N_i}\frac{1}{|N_i|}V^{l}h^{l}_j \,,
\end{equation} 
\begin{equation}\label{GCN}
h^{l+1}_i=\sigma(W^{l}h^{l}_i+z^{l}_i) \,.
\end{equation} 

We utilize graph attention to aggregate graph-level representations to make the prediction.  The graph representation is computed the same as the multiplicative attention \cite{luong2015effective}, where $h^{L}_i$ is the $i$-th node representation at the last layer, $h^c$ is the input representation $<$cls$>$, $\alpha_{i}$ is the importance of the $i$-th node, and $h^g$ is the graph representation: 

\begin{gather}
	\alpha_{i}=\frac{h^c\sigma(W_1h^{L}_i)}{\sum_{j\in N}h^c\sigma(W_1h^{L}_j)} \,, \\
	h^g=\sum_{j\in N}\alpha_{j}^{L}h^{L}_j  \,.
\end{gather}

We concatenate the input representation $h^c$ with the graph representation $h^g$ as the input of a Multi-Layer Perceptron (MLP) to compute the confidence score $score{(q,a)}$. The probability of the answer candidate $a$ to the question $a$ can be computed as follows, where $A$ is the set of candidate answers: 

\begin{equation}
p(q,a)=\frac{e^{score{(q,a)}}}{\sum_{a^{'} \in A}e^{score{(q,a^{'})}}} \,.
\end{equation}

Finally, we select the answer with the highest confidence score as the predicted answer.

\section{Experiments}
In this section, we conduct experiments to prove the effectiveness of our proposed approach. 
To dig into our approach, we perform ablation studies to explore the different effects of heterogeneous knowledge sources and graph-based reasoning models.
We study a case to show how our model can utilize the extracted evidence to get the right answer.
We also show some error cases to point directions to improve our model.

\subsection{Experiment Settings}
The CommonsenseQA \cite{talmor2019commonsenseqa} dataset contains 12,102 examples, include 9,741 for training, 1,221 for development and 1,140 for test. 

 We select XLNet large cased \cite{xlnet2019yang} as the pre-trained model. We concatenate ``The answer is'' before each choice to change each choice to a sentence. The input format for each choice is ``$<$evidence$>$ $<$sep$>$ question $<$sep$>$ The answer is $<$choice$>$ $<$cls$>$''. Totally, we get 5 confidences scores for all the choices then we adopt the softmax function to calculate the loss between the predictions and the ground truth. We adopt cross-entropy loss as our loss function. In our best model on the development dataset, we set the batch size to 4 and learning rate to 5e-6. We set max length of input to 256. We use Adam  \cite{kingma2014adam} with $\beta_1$ = 0.9, $\beta_2$ = 0.999 for optimization. We set GCN layer to 1. We train our model for 2,800 steps (about one epoch) and get the results 79.3\% on development dataset and 75.3\% on blind test dataset.
 
 \subsection{Baselines}

For the compared methods, we select models and classify them into 4 groups. \textbf{Group 1}: models without descriptions or papers, \textbf{Group 2:} models without extracted knowledge, \textbf{Group 3}: models with extracted structured knowledge and \textbf{Group 4}: models with extracted unstructured knowledge.

\begin{itemize}
    \item \textbf{Group 1}: models without description or papers. These models include SGN-lite, BECON (single), BECON (ensemble),  CSR-KG and CSR-KG (AI2 IR).
    \item \textbf{Group 2}: models without extracted knowledge, including BERT-large \cite{DevlinCLT19}, XLNet-large \cite{xlnet2019yang} and RoBERTa \cite{roberta2019liu}. These models adopt pre-trained language models to finetune on the training data and make predictions directly on the test dataset without extracted knowledge.
    \item \textbf{Group 3}: models with extracted structured knowledge, including KagNet~\cite{kag2019}, BERT + AMS \cite{align2019zhi} and BERT + CSPT. These models utilize structured knowledge ConceptNet to enhance the model to make predictions. KagNet extracts schema graphs from ConceptNet and utilize hierarchical path-based attention mechanism to infer answers.  BERT + AMS  constructs a commonsense-related multi-choice question answering dataset according to ConcepNet and pre-train on the generated dataset. BERT + CSPT first trains a generation model to generate synthetic data from ConceptNet, then finetunes RoBERTa on the synthetic data and Open Mind Common Sense (OMCS) corpus.
    \item \textbf{Group 4}: models with extracted unstructured knowledge, including CoS-E \cite{RajaniMXS19}, HyKAS, BERT + OMCS,  AristoBERTv7, DREAM, RoBERT + KE, RoBERTa + IR and RoBERTa + CSPT. Cos-E \cite{RajaniMXS19} constructs human-annotated evidence for each question and generates evidence for test data. HyKAS and BERT + OMCS models pre-train BERT whole word masking model on the OMCS corpus. AristoBERTv7 utilizes the information from machine reading comprehension data RACE \cite{LaiXLYH17} and extracts evidence from text sources such as Wikipedia, SimpleWikipedia, etc. DREAM adopts XLNet-large as the baseline and extracts evidence from Wikipedia. RoBERT + KE, RoBERTa + IR and RoBERTa + CSPT adopt RoBERTa as the baseline and utilize the evidence from Wikipedia, search engine and OMCS, respectively.
\end{itemize}

It should be noted that these methods either utilize evidence from structured or unstructured knowledge sources, failing to take advantages of both sources simultaneously. RoBERT + CSPT adopts knowledge from ConceptNet and OMCS, but the model pre-trains on the sources without explicit reasoning over the evidence, which is different from our approach.

\subsection{Experiment Results and Analysis}
\begin{table}[htbp]
    \centering
    \begin{tabular}{c|l|c|c}
    \toprule
    Group & Model & Dev Acc &Test Acc \\
    \midrule
    
   \multirowcell{4}{\textbf{Group 1}} & SGN-lite & - & 57.1 \\
    & BECON (single) & - & 57.9 \\
    & BECON (ensemble) & - & 59.6 \\
    & CSR-KG & - & 61.8 \\
    & CSR-KG (AI2 IR) & - & 65.3 \\
    
    \hline
    \hline
    
    \multirowcell{3}{\textbf{Group 2}} & BERT-large  & - & 56.7 \\
    & XLNet-large & - & 62.9 \\
    & RoBERTa(single)  & 78.5 & 72.1 \\
    & RoBERTa(ensemble) & - & 72.5 \\
    
    \hline
    \hline
    
    \multirowcell{2}{\textbf{Group 3}} &   KagNet &  - & 58.9 \\
    & BERT + AMS  & - & 62.2 \\
    & RoBERTa + CSPT & 76.2 & 69.6 \\
    
    \hline
    \hline
    
    \multirowcell{9}{\textbf{Group 4}} & Cos-E & - & 58.2 \\
    & BERT + OMCS & 68.8 & 62.5 \\
    & HyKAS & - & 62.5 \\
    & AristoBERTv7 & - & 64.6 \\
    & DREAM & 73.0 & 66.9 \\
    & RoBERT + KE & 77.5 & 68.4 \\
    & RoBERTa + CSPT & 76.2 & 69.6 \\
    & RoBERTa + IR & 78.9 & 72.1 \\
    
    \hline
    \hline
    
    & Our Model & \textbf{79.3} & \textbf{75.3} \\
    
    \bottomrule
    \end{tabular}
    \caption{Results on CommonsenseQA development and blind test dataset.}  
    \label{table:results}
\end{table}

The results on CommonsenseQA development dataset and blind test dataset are shown in Table \ref{table:results}. Our model achieves the best performance on both datasets. In the following comparisons we focus on the results on test dataset. Compared with the model in group 1, we can get more than  10\% absolute accuracy than these methods. Compared with models without extracted knowledge in group 2, our model also enjoys 2.8\% absolute gain over the strong baseline RoBERTa (ensemble). XLNet-large is our baseline model and our approach can get 12.4\% absolute improvement over the baseline and this approves the effectiveness of our approach. Compared to models with extracted structured knowledge in group 3, our model extracts graph paths from ConceptNet for graph-based reasoning rather than for pre-training, and we also extract evidence from Wikipedia plain texts, which brings 13.1\% and 5.7\% gains over BERT + AMS and ROBERTa + CSPT respectively. Group 4 contains model which utilizes unstructured knowledge such as Wikipedia or OMCS, etc. Compared with these methods, we not only utilize Wikipedia to provide unstructured evidences but also construct graphs to get the structural information. We also utilize the evidence from structure knowledge base ConceptNet. Our model achieves 3.2\% absolute improvement over the best model RoBERTa + IR in this part.

\subsection{Ablation Study}
In this section, we perform ablation studies on the development dataset\footnote{The dataset restricts to submit the results no more than every two weeks.} to dive into the effectiveness of different components in our model. 
We first explore the effect of different components in graph-based reasoning. Then we dive into the heterogeneous knowledge sources and see their effects. 

In the graph-based reasoning part, we dive into the effect of topology sort algorithm for learning contextual word representations and graph inferences with GCN and graph attention. We select XLNet + Evidence as the baseline. In the baseline, we simply concatenate all the evidence into XLNet and adopt the contextual representation for prediction. By adding topology sort, we can obtain a 1.9\% gain over the baseline. This proves that topology sort algorithm can fuse the graph structure information and change the relative position between words for better contextual word representation. The graph inference module brings 1.4\% benefit, showing that GCN can obtain proper node representations and graph attention can aggregate both word and node representations to infer answers. Finally, we add topology sort, graph inference module together to get a 3.5\% improvement, proving these models can be complementary and achieve better performance.
    
 \begin{table}[h]
     \centering
      \resizebox{0.45 \textwidth}{1.2cm}{
     \begin{tabular}{l|c}
     \toprule
        Model  &  Dev Acc \\
    \midrule
        XLNet + E & 75.8 \\
        XLNet + E + Topology Sort & 77.7 \\
        XLNet + E + Graph Inference& 77.2 \\
        XLNet + E + Topology Sort + Graph Inference & \textbf{79.3} \\
    \bottomrule
     \end{tabular}}
     \caption{Ablation studies on reasoning components in our model. E denotes evidence.}
     \label{table:model_ablation}
 \end{table}

 Then we perform ablations studies on knowledge sources to see the effectiveness of ConceptNet and Wikipedia sources. The results are shown in Table \ref{table:source_ablation}, ``None'' represents we only adopts the XLNet \cite{xlnet2019yang} large model as the baseline. When we add one knowledge source, the corresponding graph-based reasoning models are also added. From the results, we see that the structured knowledge ConceptNet can bring 6.4\% absolute improvement and the Wikipedia source can bring 4.6\% absolute improvement. This proves the benefits of ConceptNet or Wikipedia source. When combining ConceptNet and Wikipedia, we can enjoy a 9.4\% absolute gain over the baseline. This proves that heterogeneous knowledge sources can achieve better performance than single one and different sources in our model and they are complementary to each other. 
\begin{table}[h]
    \centering
    \begin{tabular}{l|c}
    \toprule
    Knowledge Sources   & Dev Acc \\
    \midrule
      None   &  68.9 \\
      ConceptNet & 75.3 \\
      Wikipedia & 73.5 \\
      ConceptNet + Wikipedia & \textbf{79.3}\\
    \bottomrule
    \end{tabular}
    \caption{Ablation studies on heterogeneous knowledge sources. ``None'' represents we only use XLNet baseline and do not utilize knowledge sources.}
    \label{table:source_ablation}
\end{table}

\subsection{Case Study}

\begin{figure}[h]
	\centering
	\includegraphics[width=0.45\textwidth]{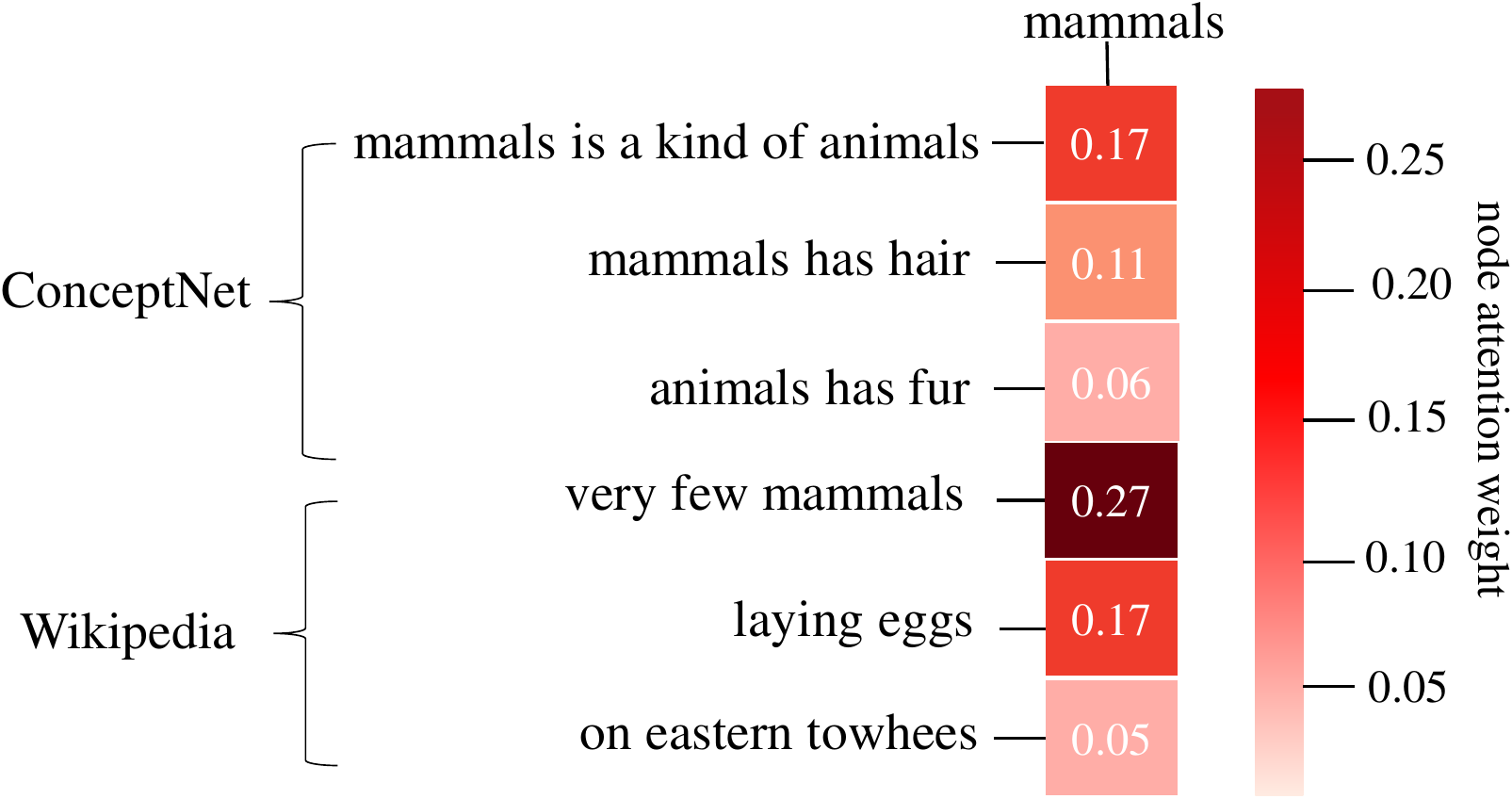}
	\caption{An attention heat-map for the question ``Animals who have hair and don't lay eggs are what?'' and the answer ``mammals''. The nodes in ConpcetNet are in natural language format and the template is: IsA (is a kind of), HasA (has).}
	\label{fig:case_study}
\end{figure}

In this section, we select a case to show that our model can utilize the heterogeneous knowledge sources to answer questions. As shown in Figure \ref{fig:case_study}, the question is ``Animals who have hair and don't lay eggs are what?'' and the answer is ``mammals''. The first three nodes are from ConceptNet evidence graph. We can see that ``mammals is animals'' and ``mammals has hair'' can provide information about the relation between ``mammals'' and two concepts ``animals'' and  ``hair''. More evidence is needed to show the relation between ``lay eggs'' and ``mammals''. The last three nodes are from Wikipedia evidence graph and they can provide the information that ``very few mammals lay eggs''. The examples also show that both sources are necessary to infer the right answer.

\subsection{Error Analysis}
\begin{figure}[h]
	\centering
	\includegraphics[width=0.45\textwidth]{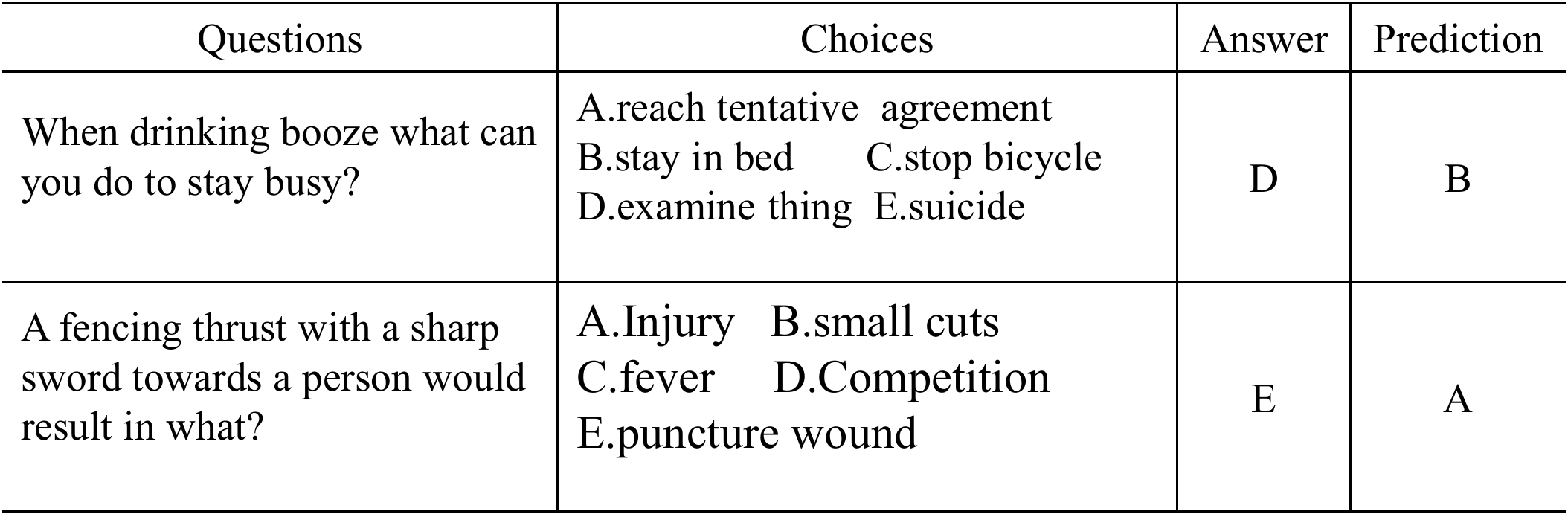}
	\caption{Error cases of our model on the development dataset.}
	\label{fig:error_case}
\end{figure}

We randomly select 50 error examples from the development dataset and the reasons are classified into three categories: the lack of evidence, similar evidence and dataset noise. There are 10 examples which are lack of evidence.
For example, the first example in Figure \ref{fig:error_case} extracts no triples from ConceptNet and the evidence from Wikipedia does not contain enough information to get the right answer. 
This problem can be alleviated by utilizing more advanced extraction strategies and adding more knowledge sources. There are 38 examples which extract enough evidence but the evidence are too similar to distinguish between choices.
For example, the second example in Figure \ref{fig:error_case} has two choices ``injury'' and ``puncture wound'', the evidence from both sources provides similar information.
More evidence from other knowledge sources is needed to alleviate this problem. We also find there are 2 error examples which have 2 same choices\footnote{example id: e5ad2184e37ae88b2bf46bf6bc0ed2f4, fa1f17ca535c7e875f4f58510dc2f430}.

\section{Related Work}
\textbf{Commonsense Reasoning} Commonsense reasoning is a challenging direction since it requires reasoning over external knowledge beside the inputs to predict the right answer. Various downstream tasks have been released to address this problem like ATOMIC\cite{sap2019atomic}, Event2Mind\cite{rashkin2018event2mind}, MCScript 2.0\cite{ostermann2019mcscript2}, SWAG\cite{ZellersBSC18}, HellaSWAG\cite{ZellersHBFC19} and Story Cloze Test\cite{MostafazadehCHP16}.
 
Recently proposed CommonsenseQA\cite{talmor2019commonsenseqa} dataset derived from ConceptNet\cite{SpeerCH17} and the choices have the same relation with the concept in the question. 
Recently, \citet{RajaniMXS19} explores adding human-written explanations to solve the problem. 
\citet{kag2019} extracts evidence from ConceptNet to study this problem. This paper focuses on automatically extracting evidence from heterogeneous external knowledge and reasoning over the extracted evidence to study this problem.

\textbf{Knowledge Transfer in NLP} Transfer learning has played a vital role in the NLP community. 
Pre-trained language models from large-scale unstructured data like ELMo \cite{PetersNIGCLZ18}, GPT \cite{radford2018improving}, BERT \cite{DevlinCLT19}, XLNet \cite{xlnet2019yang}, RoBERTa \cite{roberta2019liu} have achieved significant improvements on many tasks.
This paper utilizes XLNet \cite{xlnet2019yang} as the backend and propose our approach to study the commonsense question answering problem.

\textbf{Graph Neural Networks for NLP} Recently, Graph Neural Networks (GNN) has been utilized widely in NLP. For example,  \citet{SunGWGJLSD19} utilizes Graph Convolutional Networks (GCN) to jointly extract entity and relation. \citet{zhang2018graph} applies GNN to relation extraction over pruned dependency trees and achieves remarkable improvements. 
GNN has also been applied into muli-hop reading comprehension tasks \cite{TuWHTHZ19,KunduKSC19,JiangJCB19}. This paper utilizes GCN to represent graph nodes by utilizing the graph structure information, followed by graph attention which aggregates the graph representations to make the prediction.

\section{Conclusion}
In this work, we focus on commonsense question answering task and select CommonsenseQA \cite{talmor2019commonsenseqa} dataset as the testbed. 
We propose an approach consisting of knowledge extraction and graph-based reasoning. In the knowledge extraction part, we extract evidence from heterogeneous external knowledge including structured knowledge source ConceptNet and  Wikipedia plain texts. 
In the graph-based reasoning part, we propose a graph-based  approach  consisting  of  graph-based  contextual  word representation  learning  module  and  graph-based  inference module to select the right answer. Results show that our model achieves state-of-the-art on CommonsenseQA\cite{talmor2019commonsenseqa} dataset.


\section{Acknowledgement}
Songlin Hu is the corresponding author. We thank the anonymous reviewers for providing valuable suggestions.

\bibliography{scholar.bib}
\bibliographystyle{aaai}

\end{document}